\documentclass[a4paper]{svproc}

\usepackage{graphicx}        
\usepackage[bottom]{footmisc} 
\usepackage{multicol}        
\usepackage{url}             

\usepackage{xcolor}
\usepackage[
    colorlinks=true,
    linkcolor=black,
    citecolor=black,
    urlcolor=blue
]{hyperref}
\usepackage{CJKutf8}
\definecolor{darkgreen}{rgb}{0.0, 0.65, 0.0}
\definecolor{awesome}{rgb}{1.0, 0.13, 0.32}

\usepackage{amsmath,amssymb}
\usepackage{booktabs}
\usepackage[normalem]{ulem}

\graphicspath{{figures/}}

\usepackage{xcolor}
\usepackage{threeparttable}
\usepackage{booktabs}
\usepackage{multirow}

\newcommand{\limebox}[1]{\setlength{\fboxsep}{0pt}{\colorbox{lime}{#1}}}
\newcommand{\pinkbox}[1]{\setlength{\fboxsep}{0pt}{\colorbox{yellow}{#1}}}

\definecolor{darkorange}{rgb}{1.0, 0.55, 0.0}

\newcommand{\rev}[1]{\begingroup\color{magenta}#1\endgroup}

\begin{document}
\mainmatter                  


\title{Extreme Motion Generation \\via Hybrid Null-Space Control \\for Straight-Line Path Following}
\titlerunning{Extreme Motion Generation}

\author{Xinyi Yuan
        \and Weiwei Wan$^{*}$
        \and Kensuke Harada}
\authorrunning{Yuan et al.}
\tocauthor{Xinyi Yuan,Weiwei Wan,Kensuke Harada}

\institute{Graduate School of Engineering Science, The University of Osaka, Japan.}

\maketitle

\begin{abstract}
This work studies \textit{extreme motion generation}, which aims to maximize the Cartesian path length along a pre-defined trajectory within the manipulator's workspace. This objective is important in industry as long as path-following is fundamental to a large variety of tasks such as surface coating and welding. More critically, extreme motion enables a fixed-base manipulator to exploit the kinematic capability under limited reachability. However, such exploitation is challenging in practice, as the manipulator must actively avoid the safety boundary through execution, which is inherently a long-horizon problem. Accordingly, we claim that long-horizon decision-making should be delegated to a learning-based policy to maximize exploitation, while a classical model-based controller covers the near-boundary region, where the learning policy degrades sharply due to sparse data coverage. In detail, our proposed method is a step-level hybrid controller that switches between an RL-based and a model-based controller according to the normalized joint-limit distance. The initial joint configuration is sampled through conditional diffusion-based sampling, which improves the achievable path length based on the learned motion prior. We evaluate the proposed framework on 10,000 straight-line path-following tasks with a 7-DoF Franka FR3, extending the average rollout length by 27\% over the model-based baseline. Notably, certain tasks yield a pronounced extension toward the motion extreme, as reflected in the maximum improvement reported in the statistical results. The project website and related videos of this paper can be found at \url{https://yuan-xinyi.github.io/extreme-motion-generation/}.

\end{abstract}

\section{Introduction}
Consider a compact, fixed-base manipulator executing a coating or welding operation over a large work surface. Given a starting position and a motion direction, intuition would suggest that the maximum reachable point is determined by the reachable workspace. However, the motion is continuous and strictly constrained by safety boundaries such as joint limits and tool-orientation tolerance along each waypoint. Therefore, the practical bottleneck is that one of those safety constraints is violated and causes termination long before the end-effector approaches the edge of the reachable workspace. Consequently, the practically achievable path length is influenced by how long the safety boundaries can be deferred from being met during the motion execution process. We refer to this problem as \textit{extreme motion generation} that maximizes the Cartesian path length along a pre-defined trajectory within the manipulator's workspace. This is particularly important in industrial automation, where the length of a single continuous motion directly limits the area a manipulator can process in one setup. Rather than scaling up the manipulator or repeatedly relocating the base, simply exploiting the extreme motion offers a solution to enlarge the effective working range.

However, extreme motion generation, especially under precise path-following, is challenging in practice. The joint motion at the current step may gradually steer the subsequent motion toward a safety boundary, ultimately forcing it to terminate. Therefore, the problem is inherently a long-term decision-making problem, as each step must account for its consequences over the entire motion. Unfortunately, current motion generation tools have certain drawbacks with respect to the requirement of both robustness and exploitation. On the one hand, a classical controller optimizes an instantaneous secondary objective while realizing the desired end-effector motion. Such controllers are myopic, optimizing an instantaneous surrogate and overlooking its long-horizon consequences. Yet near the joint limits, their behavior is robust and predictable by design. On the other hand, the learning-based policy is designed to have the foresight suitable for exploitation. However, the learning paradigm is based on probabilistic distribution approximation, and near-boundary configurations are rarely sampled. As a result, the derived policy is particularly fragile near the boundary regions that are most critical to extreme motion generation.

Motivated by this complementarity, we delegate the long-horizon decision-making in the well-sampled interior to a Reinforcement-Learning (RL) policy, where foresight can be trusted. The near-boundary region is handled by the classical controller, whose robustness is guaranteed by construction. The two regimes are coordinated at every control step by a hysteresis rule on the normalized joint-limit distance, ensuring that each model operates within the region where it is most reliable. In addition, a conditional diffusion model initializes the manipulator on a robust starting configuration prior to execution of the hybrid controller.

We evaluate the proposed framework on a large set of straight-line path-following tasks for a 7-DoF Franka FR3, measuring performance as the achieved path length relative to a per-task reference length. The contribution of each component is then analyzed through the seed-by-controller protocol. For the initial joint configuration, the diffusion-based selection is compared against several hand-designed heuristic seeds; for the controller, the hybrid framework is compared against the classical and the learning-based controllers in isolation. We further conduct ablation studies on the guidance scale, the number of diffusion sampling steps, and the hysteresis switching thresholds, providing practical insights into the effect of each design choice.

\section{Related Work}

\subsection{Manipulability and Reachability}
Extreme motion generation is essentially related to the manipulator's reachable workspace and local motion dexterity during motion execution. Manipulability and reachability describe local instantaneous motion dexterity and global spatial feasibility, respectively.

The manipulability ellipsoid \cite{yoshikawa1985manipulability} characterizes the relationship between joint and task-space velocities via the Jacobian, and its projection along a task direction quantifies directional capability. To exploit this measure online, Jin et al.~\cite{jin2017manipulability} adopted a dynamic neural network for inverse-free manipulability optimization. Recent work integrates manipulability into planning: Shen et al.~\cite{shen2023adaptive} proposed a manipulability-aware $\text{RRT}^{\star}$ that embeds the measure into both cost and step size, and Cai et al.~\cite{cai2025just} balanced path cost against manipulability to mitigate singularities during planning. The above studies can effectively keep the manipulator away from singular configurations but act only on the instantaneous state and overlook the long-horizon consequences of the current motion.

Complementary to local dexterity, reachability analysis delineates the global spatial limit that bounds how far any motion can extend. Early studies established geometric \cite{yang2009placement} and voxel-based \cite{makhal2018reuleaux} reachability maps to automate base placement without exhaustive IK. Moving toward dynamic settings, Feng et al.~\cite{feng2025predictive} introduced predictive reachability via world models for embodiment selection, while Jauhri et al.~\cite{jauhri2022robot} used reachability priors to accelerate RL-based control, and Selim et al.~\cite{selim2022safe} proposed a reachability-based safety layer with formal guarantees. These methods characterized the reachable space with a static, precomputed envelope. However, they do not address how to actively defer the safety boundaries during execution so that the end-effector approaches this envelope as closely as possible. Bridging this gap is precisely the goal of extreme motion generation.

\subsection{Learning-based Motion Generation}
In contrast to optimizing an instantaneous objective, learning-based methods capture the multi-modal distribution of feasible configurations and enable long-horizon motion generation. For example, Reinforcement Learning (RL) has been applied to redundancy resolution \cite{malik2022deep}, and generative solvers such as IKFlow \cite{ames2022ikflow} and IKDiffuser \cite{zhang2025ikdiffuser} modeled the full solution manifold rather than a single solution. The related study DiffusionSeeder \cite{huang2024diffusionseeder} utilized generative models to seed downstream motion to produce multi-modal trajectory initializations, and Yoon et al.~\cite{yoon2023learning} proposed an RL-based initialization generator.

Such learned priors supply the foresight that manipulability- and reachability-based methods lack. However, the data used to train these policies is sparsely distributed near the kinematic boundaries. Consequently, the resulting policies are most fragile in exactly the near-boundary region that is most critical to extreme motion generation. This complementarity between learned foresight in the well-sampled region and classical robustness near the boundary motivates the proposed hybrid framework.

\subsection{Combining Learning- and Model-based Control}
To address the aforementioned issues, the solution could be to combine a learned policy with a model-based controller. One line of work combines the output actions additively: residual RL keeps a fixed classical controller and learns a corrective action added on top of that \cite{johannink2019residual,silver2018residual}. Another line of research switches between different controllers by safety criteria or pre-defined regions. For example, Mehmood et al.~\cite{mehmood2022black} switched to a verified backup controller under runtime reachability checks. Similarly, Alshiekh et al.~\cite{alshiekh2018safe} proposed to override unsafe learned actions whenever the agent approaches the boundary. Likewise, Kurtz et al.~\cite{kurtz2021control} continuously filtered the nominal action to keep it within a safe set, as in control-barrier-function safety filters for manipulators.

Generally, our framework follows the switching rather than the additive paradigm. Similar to the safety-driven methods above, we trigger the backup classical controller near the safety boundary. However, our motivation is not merely to stay safe but to keep advancing the motion toward its kinematic limit, which is the core objective of extreme motion generation. Accordingly, we partition the configuration space by proximity to the joint-limit boundary and assign each region to the controller most reliable there.

\section{Problem Formulation}
\begin{figure}[!htbp]
    \centering
    \includegraphics[width=\linewidth]{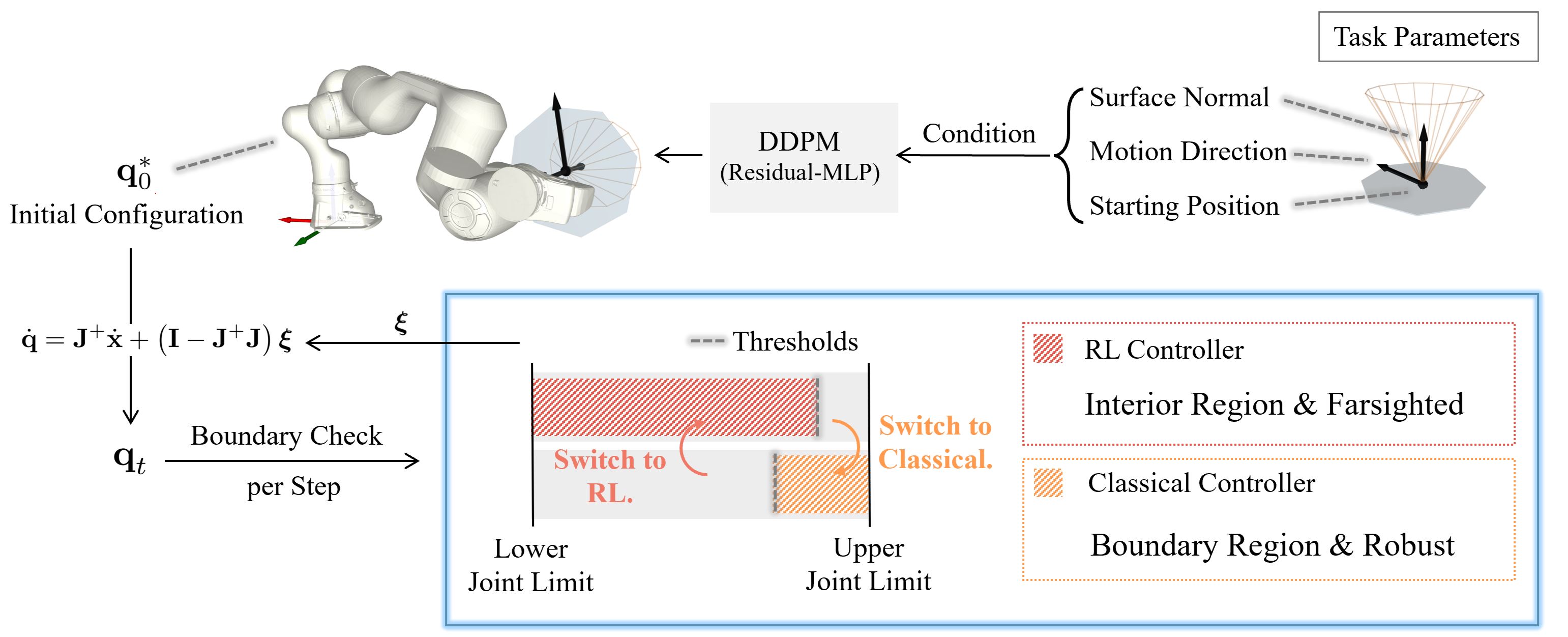}
    \caption{Overview of the two-stage framework for extreme motion generation.}
    \label{overview}
\end{figure}

While extreme motion generation applies broadly to constrained path following, we here specify it as extreme straight-line path following. Specifically, a fixed-base manipulator advances its end-effector along the line with a tool-orientation constraint. As illustrated in Fig.~\ref{overview}, solving this problem requires first determining the initial joint configuration from which the motion starts and then generating the motion that maximizes the achievable path length.

We formalize the above task as follows. Consider a manipulator with $n$ degrees of freedom, where $\mathbf{q} \in \mathbb{R}^{n}$ denotes the joint configuration. Let $\mathbf{p}(\mathbf{q}) \in \mathbb{R}^{3}$ denote the end-effector position and $\mathbf{z}(\mathbf{q}) \in \mathbb{R}^{3}$ denote the TCP local $z$-axis. A straight-line path-following task is specified by the condition vector:
\begin{equation}
\label{cond}
\mathbf{c} = [\mathbf{p}_{0}, \mathbf{d}, \mathbf{n}] \in \mathbb{R}^{9},
\end{equation}
where $\mathbf{p}_{0}$ denotes the starting position, and $\mathbf{d}$ and $\mathbf{n}$ are unit vectors representing the desired motion direction and task-plane normal, respectively. 

Starting from an initial configuration $\mathbf{q}_{0}^{*}$, the policy $\pi$ determines the joint motion at every control step to maximize the path-following length, producing a joint trajectory $\{\mathbf{q}_{t}^{*}\}_{t=0}^{T}$ initialized at $\mathbf{q}_{0}^{*}$, where $T$ is the termination step at which any safety constraint is first violated. This task can be formalized as a constrained optimization problem as:
\begin{align}
\max_{\mathbf{q}_{0}^{*},\,\pi} \quad & \mathcal{L}(\mathbf{q}^{*}_{0},\,\pi;\,\mathbf{p}_{0}, \mathbf{d}, \mathbf{n}) \\
\text{s.t.} \quad & \|\mathbf{p}(\mathbf{q}_{t}^{*}) - \mathbf{p}_{t}\| \leq \epsilon_{\text{pos}}, \\
& \arccos\!\left( \mathbf{z}(\mathbf{q}_{t}^{*})^{\top} \mathbf{n} \right) \leq \theta_{\max}, \label{eq:ori_constraint} \\
& \mathbf{q}_{\min} \preceq \mathbf{q}_{t}^{*} \preceq \mathbf{q}_{\max}, \label{eq:jl_constraint}
\end{align}
where $\mathcal{L}$ denotes the path-following length, $\mathbf{q}_0^*$ and $\pi$ are the initial configuration and the control policy to be optimized, and $(\mathbf{p}_0, \mathbf{d}, \mathbf{n})$ specify the task conditions as above. The $\epsilon_{\text{pos}}$, $\theta_{\max}$, and $[\mathbf{q}_{\min}, \mathbf{q}_{\max}]$ are the position tolerance, orientation tolerance, and joint limits, respectively.

During this process, since only the $3$-DoF end-effector position must be precisely tracked, the remaining freedom makes the manipulator kinematically redundant for $n>3$. This redundancy is resolved in the null space of the position Jacobian, which decomposes the commanded joint velocity into a task term and a null-space term,
\begin{equation}
\label{eq:nullspace}
\dot{\mathbf{q}} = \mathbf{J}^{+}\dot{\mathbf{x}} + (\mathbf{I}-\mathbf{J}^{+}\mathbf{J})\,\boldsymbol{\xi},
\end{equation}
where $\mathbf{J}^{+}$ is the damped pseudo-inverse and $(\mathbf{I}-\mathbf{J}^{+}\mathbf{J})$ projects the secondary command $\boldsymbol{\xi}$ onto the null space. Within this null space, several secondary objectives can be pursued without disturbing the end-effector position:
\begin{equation}
\label{eq:secondary}
\boldsymbol{\xi}=k_{\mu}\,\nabla_{\mathbf{q}}w_{\mathbf{d}}(\mathbf{q})\;-\;k_{\mathrm{jl}}\,\frac{\mathbf{q}-\mathbf{q}_{\mathrm{mid}}}{\mathbf{q}_{\max}-\mathbf{q}_{\min}}\;+\;k_{\theta}\,g(\theta)\,\nabla_{\mathbf{q}}\!\big(\mathbf{z}(\mathbf{q})^{\top}\mathbf{n}\big),
\end{equation}
where the secondary command $\boldsymbol{\xi}$ combines a directional-manipulability term $w_{\mathbf{d}}$ that retreats from singularities along $\mathbf{d}$, a centering term that pulls each joint toward the middle $\mathbf{q}_{\mathrm{mid}}$ of its range, and a cone term gated by $g(\theta)$ that activates only as the tool axis approaches the $\theta_{\max}$ boundary. All subsequent controllers share this task decomposition, differing only in the strategy used to generate the secondary command $\boldsymbol{\xi}$.

\section{Extreme Motion Generation}
Following the two-stage framework illustrated in Fig.~\ref{overview}, this section provides detailed explanations of each stage. Before execution ($t=0$), a conditional diffusion model samples the initial configuration $\mathbf{q}_{0}^{*}$ from the task condition $\mathbf{c}$. During execution ($t>0$), a hybrid controller generates the motion online, switching between an RL controller in the interior and the classical controller near the joint-limit boundary.


\subsection{Initial Joint Configuration Selection}

\begin{figure}[!htbp]
    \centering
    \includegraphics[width=\linewidth]{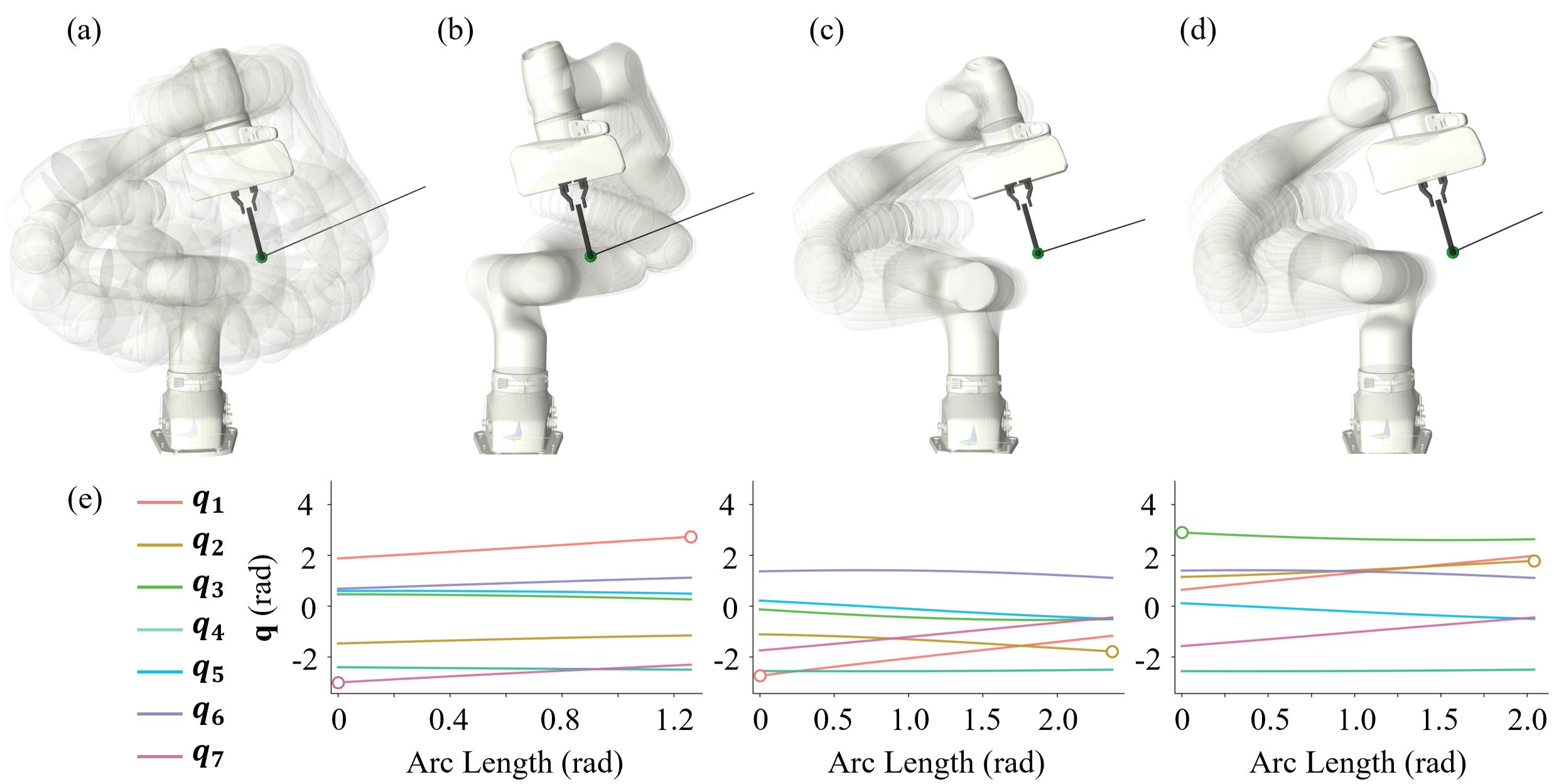}
    \caption{Self-motion manifold for a fixed end-effector pose. (a) The continuous manifold without joint-limit constraints; (b)--(d) Configurations from the disconnected branches induced by the joint limits, sharing the same pose but differing sharply in joint space and reachable path length. (e) Joint trajectories of the three branches in b--d along the arc length, with circles marking the joint at its limit: branch~b bounded by $\mathbf{q}_{1}$ and $\mathbf{q}_{7}$, branch~c by $\mathbf{q}_{1}$, and $\mathbf{q}_{2}$ branch~d by $q_{2}$ and $\mathbf{q}_{3}$.}
    \label{smm}
\end{figure}

For many tasks, the achievable path length is largely determined by the initial joint configuration even before motion begins. For example, for a 7-DoF arm, a fixed end-effector pose constrains 6-DoF, and the remaining one forms a one-dimensional self-motion manifold (SMM). However, this does not imply that the arm can move freely along the manifold, as the joint limits split it into several disconnected branches. As shown in Fig.~\ref{smm}, configurations on different branches share the same pose but differ sharply in joint space and in reachable path length.

Therefore, this selection module aims to provide the hybrid controller with an initial joint configuration within a favorable branch. Consequently, the learned motion prior should incorporate configurations from various branches that yield a sufficient path length. For each task, we enumerate configurations across branches by cone-constrained IK solved by IKSel \cite{yuan2026iksel}, score each by a clean rollout under the classical controller, and retain the top-$K'$ scoring candidates ($K'{=}6$) that remain length-stable under small task perturbations as training labels.

The distribution of these initial joint configurations is approximated with a conditional DDPM~\cite{ho2020denoising}, where a residual-MLP denoiser generates a $7$-dimensional joint configuration $\mathbf{q}_{0}$ conditioned on the task vector $\mathbf{c}=[\mathbf{p}_{0};\mathbf{d};\mathbf{n}]$, trained under Classifier-Free Guidance (CFG) \cite{ho2022classifier} by dropping the condition with probability $p_{\mathrm{drop}}$. At inference, the guidance scale $w$ in
\begin{equation}
\label{eq:cfg}
\mathbf{v} = \mathbf{v}_{\theta}(\mathbf{x}_{t},t,\varnothing) + w\big(\mathbf{v}_{\theta}(\mathbf{x}_{t},t,\mathbf{c}) - \mathbf{v}_{\theta}(\mathbf{x}_{t},t,\varnothing)\big)
\end{equation}
controls the task signal, where $w=1$ recovers conditional sampling and $w>1$ amplifies it. Each sampled candidate is then projected onto the exact $(\mathbf{p}_{0},\mathbf{n})$ pose by Newton's method, and the first IK-valid one is retained as the seed $\mathbf{q}_{0}^{*}$, falling back to $\mathbf{q}_{0}^{\mathrm{seed}}$ (the configuration that generated the task) if none succeeds after maximum 32 trials.

\subsection{RL-based Null-Space Controller}
The interior policy $\pi$ is trained with Proximal Policy Optimization (PPO). The observation $\mathbf{o}_{t}\in\mathbb{R}^{31}$ exposes only the quantities needed to resolve the redundancy:
\begin{equation}
\label{eq:obs}
\mathbf{o}_{t}=\big[\,\bar{\mathbf{q}};\ \bar{\mathbf{q}}^{\odot 2};\ \mathbf{d};\ \mathbf{n};\ \mathbf{z}(\mathbf{q});\ \mathbf{z}(\mathbf{q})^{\top}\mathbf{n};\ \mathbf{z}(\mathbf{q})\times\mathbf{n};\ \mathbf{a}_{t-1}\,\big],
\end{equation}
where $\bar{\mathbf{q}}$ is the joint configuration normalized to $[-1,1]$, whose element-wise square $\bar{\mathbf{q}}^{\odot 2}$ makes the proximity to either limit directly visible, and the tool-axis terms $\mathbf{z}(\mathbf{q})$, $\mathbf{z}(\mathbf{q})^{\top}\mathbf{n}$, and $\mathbf{z}(\mathbf{q})\times\mathbf{n}$ encode the orientation error relative to the cone. The end-effector position is deliberately excluded, as lateral tracking is handled by the controller rather than the policy.

The policy outputs a $4$-dimensional action $\mathbf{a}\in[-1,1]^{4}$ whose entries are coordinates in a task-aligned null-space basis $\mathbf{B}(\mathbf{q})\in\mathbb{R}^{7\times 4}$. The basis is constructed by Gram--Schmidt: the three projected gradients of the redundancy objectives in Eq.~\eqref{eq:secondary} are orthonormalized first, and the remaining one-dimensional orthogonal complement within the $4$-dimensional null space is appended to complete the basis. The resulting command $\dot{\mathbf{q}}_{\mathrm{null}}=\mathbf{B}(\mathbf{q})\,(a_{\max}\mathbf{a})$ is added to the same task term as the classical controller in Eq.~\eqref{eq:nullspace}, so that the two regimes share one parameterization and are directly interchangeable at switching time.

Since the objective is simply to maximize the path length, the reward needs no elaborate shaping and is purely progress-based:
\begin{equation}
\label{eq:reward}
r_{t} = w_{p}\,\mathrm{clip}\!\left(\frac{(\mathbf{p}_{t+1}-\mathbf{p}_{t})^{\top}\mathbf{d}}{v\,\Delta t},\,0,\,1\right)\in[0,\,w_{p}],
\end{equation}
the per-step advance along $\mathbf{d}$ normalized by its maximum attainable value. With no terminal penalty, any violation simply ends the episode, so the return reduces to the survival time, and the policy is rewarded solely for sustaining the motion.

\subsection{Null-Space Policies Coordination}\label{sec:coordination}
The two controllers are combined at the level of individual control steps. We measure proximity to the joint-limit boundary by the normalized distance
\begin{equation}
\label{eq:rho}
\rho(\mathbf{q}) = \max_{i}\,\frac{|q_{i}-q_{\mathrm{mid},i}|}{(q_{\max,i}-q_{\min,i})/2}\in[0,1],
\end{equation}
which is $0$ at the center of the joint range and $1$ at a limit. The RL policy governs the interior and the classical controller the near-boundary shell, with switching following a hysteresis rule with two thresholds $\tau_{\mathrm{enter}}\ge\tau_{\mathrm{exit}}$ to reduce chattering: while the RL policy is active, control passes to the classical controller once $\rho\ge\tau_{\mathrm{enter}}$, and returns to the RL policy only after $\rho$ falls back below $\tau_{\mathrm{exit}}$. An episode whose seed already lies in the boundary shell, $\rho(\mathbf{q}_{0}^{*})\ge\tau_{\mathrm{enter}}$, begins under the classical controller.

\section{Experiments and Analysis}
\rev{\label{sec:exp}}
All experiments were conducted using the \textit{One} robotic system\footnote{https://github.com/wanweiwei07/one}, on the 7-DoF Franka Research 3 (FR3) with a pen-shaped tool attached to the end-effector. The evaluation set comprises $10{,}000$ tasks, each specified by a starting position $\mathbf{p}_0$, a desired line direction $\mathbf{d}$, and a task-plane normal $\mathbf{n}$, sampled within the FR3 work range. These tasks are drawn from the same population as the diffusion training data, with $10\%$ held out as unseen during training to assess generalization.

Unless otherwise specified, experiments use the default setting: (1) a diffusion-based initial configuration generator using DDIM with 50 denoising steps and CFG scale $w=1.5$; (2)the diffusion-based initialization runs at most 32 trials, or falls back to the predefined seed $\mathbf{q}_0^{\mathrm{seed}}$ if none succeeds; (3) hybrid switching thresholds $\tau_{\mathrm{enter}},\tau_{\mathrm{exit}}$ of $0.98,0.94$; (4) classical secondary-objective weights $k_{\mu},k_{\mathrm{jl}},k_{\theta}$ of $0.8,0.4,0.2$.

For each evaluation task, we generate seed candidates following the procedure of Section~4.1 (up to $6$ seeds per task) and roll out every candidate under the deployed hybrid controller. We denote the longest rollout length as the \emph{reference length} ($\ell^{\mathrm{ref}}$) and utilize it for subsequent percentage comparison.

\subsection{Results Analysis}
\subsubsection{Evaluation Set Analysis}
\begin{figure}[!htbp]
    \centering
    \includegraphics[width=\linewidth]{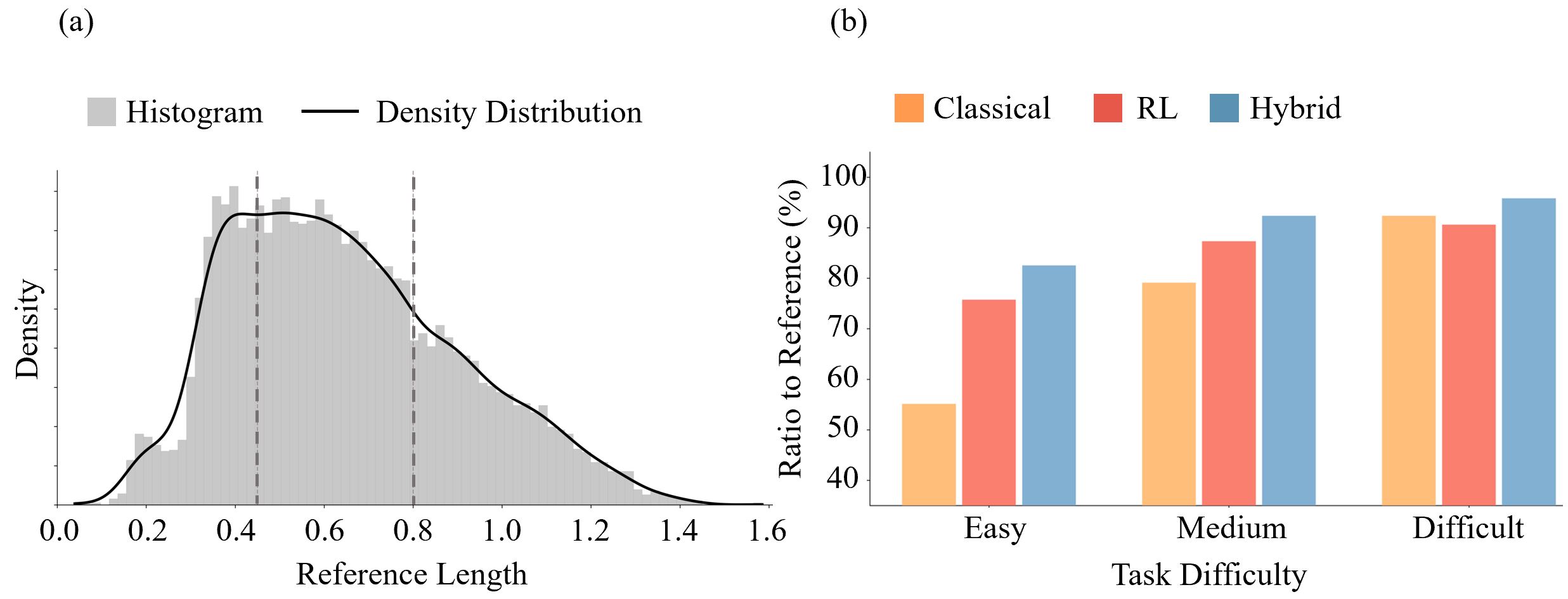}
    \caption{(a) Distribution of the reference length $\ell^{\mathrm{ref}}$ over the $10{,}000$ evaluation tasks. The dashed grey line marks the task-difficulty threshold.}
    \label{fig:eval_dist}
\end{figure}

We partition the evaluation set into difficulty levels by the absolute reference length $\ell^{\mathrm{ref}}$: Easy ($\ell^{\mathrm{ref}}\ge 0.80$\,m, $n{=}2642$), Medium ($0.45\le\ell^{\mathrm{ref}}<0.80$\,m, $n{=}4795$), and Difficult ($\ell^{\mathrm{ref}}<0.45$\,m, $n{=}2563$). Intuitively, harder tasks exhibit a shorter best achievable path, reflected in a smaller reference length. Fig. \ref{fig:eval_dist} (a) shows the distribution of $\ell^{\mathrm{ref}}$ of the complete evaluation set.

\begin{figure}[!htbp]
    \centering
    \includegraphics[width=\linewidth]{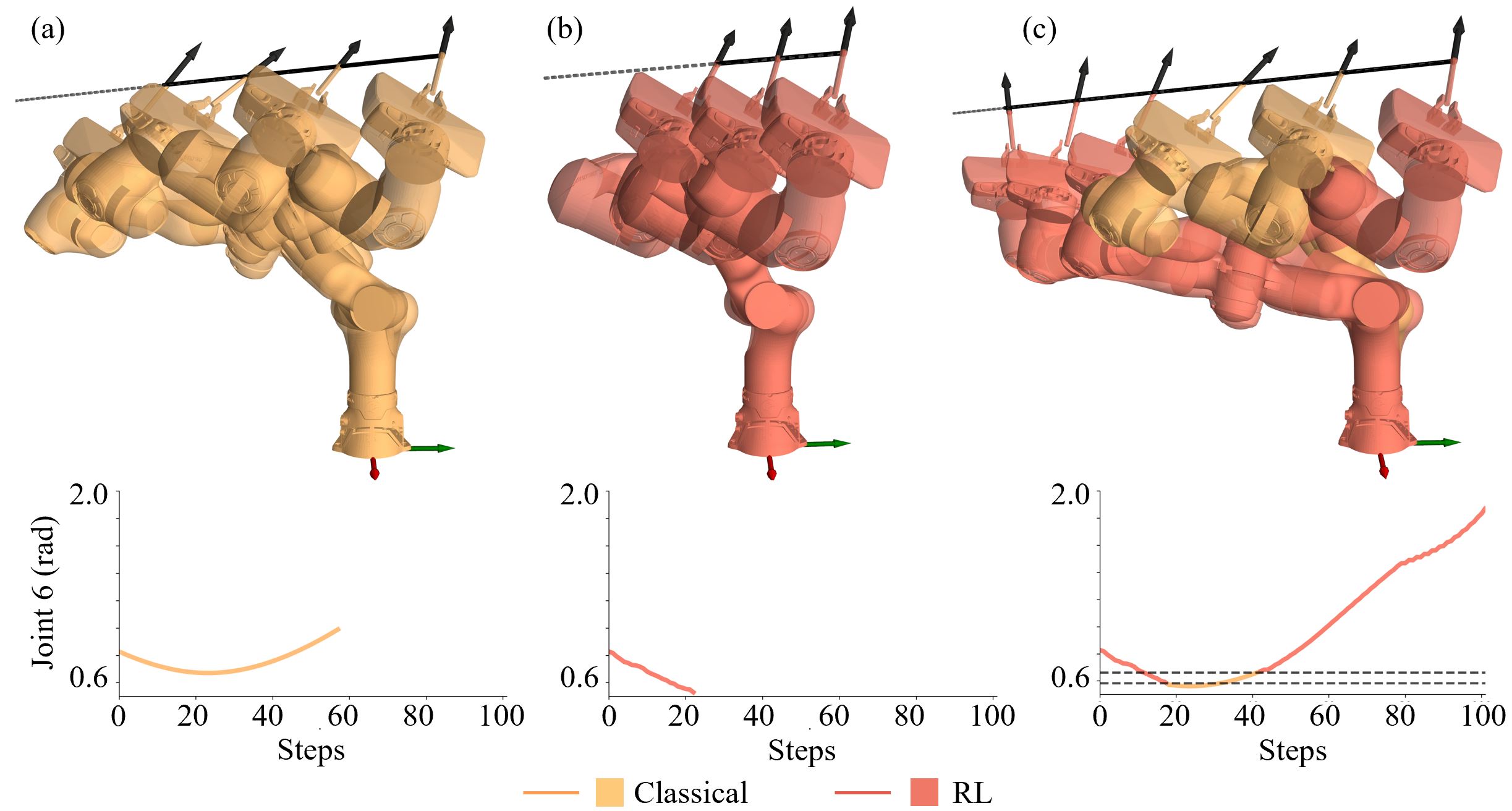}
    \caption{Qualitative comparison on a representative task: executed motion (top) and joint~6 trajectory (bottom), with color denoting the active controller. The dashed grey lines mark the $\tau_{\mathrm{enter}}/\tau_{\mathrm{exit}}$ thresholds mapped to the joint~6 limits.}
    \label{fig:main_result}
\end{figure}

\subsubsection{System Integration}
Table~\ref{tab:lnetresult} reports the path-following length $l$ (m) and its per-task ratio to the reference length $\ell^{\mathrm{ref}}$ for the three controllers under diffusion-prior seed (denoted as DP), with results bucketed by task difficulty as defined above. The occasional ratios exceeding $100\%$ arise where the DP seed reaches a more favorable branch than any of the reference seeds. The results further verify that the foresight advantage of RL diminishes as the interior margin shrinks with task difficulty: RL outperforms on Easy and Medium tasks where the margin remains, but matches Classical on Difficult ones near the boundary. Across all buckets, Hybrid attains the best of the three, recovering the largest fraction of the reference length. Fig.~\ref{fig:main_result} provides a representative illustration.

\begin{table*}[!htbp]
\centering
\setlength{\tabcolsep}{6pt}
\caption{Path-Following Performance of the Three Controllers across Difficulty Buckets on 10,000 Evaluation Tasks.}
\label{tab:lnetresult}
\begin{threeparttable}
\begin{tabular}{lllcc}
\toprule
Eval-Set & $\textbf{q}_{0}^*$ & $\pi$ 
& \shortstack{Path Length (m)\\ {\scriptsize Mean / Std / Min / Max}} 
& \shortstack{Ratio to Reference ($\%$)\\ {\scriptsize Mean / Std / Min / Max}} \\
\midrule
\multirow{3}{*}{\shortstack{All\\{\scriptsize ($n{=}10000$)}}}
& DP & Classical & \pinkbox{0.45} / 0.18 / 0.01 / 1.25 & \pinkbox{76.1} / 26.9 / 0.8 / \phantom{0}601.3 \\
& DP & RL        & 0.53 / 0.26 / 0.01 / 1.45 & 85.0 / 28.8 / 0.8 / 1017.6 \\
& DP & Hybrid    & \limebox{0.57} / 0.25 / 0.01 / 1.45 & \limebox{90.6} / 23.1 / 0.8 / 1000.8 \\
\midrule
\multirow{3}{*}{\shortstack{Easy\\{\scriptsize ($n{=}2642$)}}}
& DP & Classical & \pinkbox{0.53} / 0.25 / 0.01 / 1.25 & \pinkbox{55.0} / 27.3 / 0.8 / \phantom{0}111.1 \\
& DP & RL        & 0.74 / 0.32 / 0.01 / 1.45 & 75.7 / 31.4 / 0.8 / \phantom{0}143.5 \\
& DP & Hybrid    & \limebox{0.81} / 0.28 / 0.01 / 1.45 & \limebox{82.4} / 26.2 / 0.8 / \phantom{0}143.5 \\
\midrule
\multirow{3}{*}{\shortstack{Medium\\{\scriptsize ($n{=}4795$)}}}
& DP & Classical & \pinkbox{0.48} / 0.13 / 0.01 / 1.11 & \pinkbox{79.1} / 19.6 / 1.3 / \phantom{0}176.6 \\
& DP & RL        & 0.53 / 0.17 / 0.01 / 1.26 & 87.2 / 24.9 / 1.2 / \phantom{0}243.2 \\
& DP & Hybrid    & \limebox{0.57} / 0.14 / 0.01 / 1.26 & \limebox{92.3} / 16.9 / 1.2 / \phantom{0}235.3 \\
\midrule
\multirow{3}{*}{\shortstack{Difficult\\{\scriptsize ($n{=}2563$)}}}
& DP & Classical & \pinkbox{0.32} / 0.08 / 0.01 / 1.14 & 92.3 / 24.4 / 2.3 / \phantom{0}601.3 \\
& DP & RL        & \pinkbox{0.32} / 0.11 / 0.01 / 1.28 & \pinkbox{90.5} / 30.5 / 2.2 / 1017.6 \\
& DP & Hybrid    & \limebox{0.33} / 0.09 / 0.01 / 1.28 & \limebox{95.7} / 27.2 / 2.4 / 1000.8 \\
\bottomrule
\end{tabular}
\end{threeparttable}
\end{table*}

\subsubsection{Initial Configuration Selection}
\begin{figure}[!htbp]
    \centering
    \includegraphics[width=\linewidth]{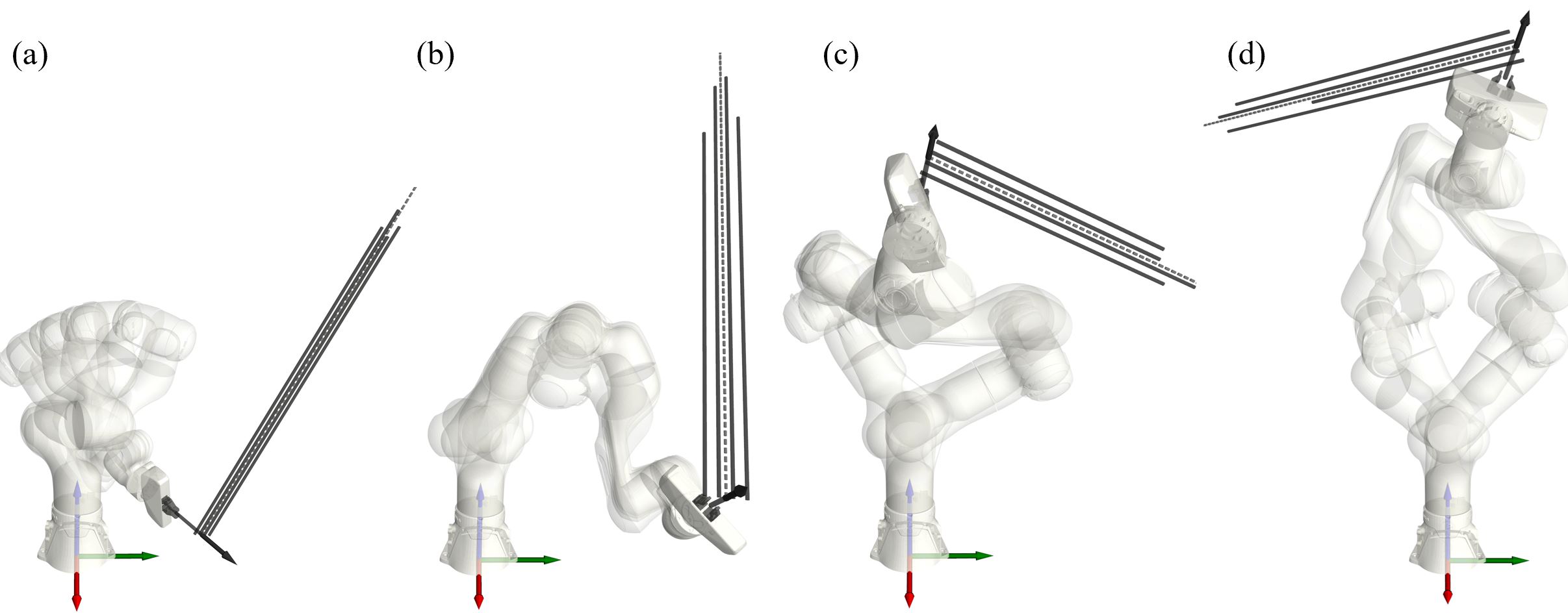}
    \caption{Multi-modal initial configurations sampled by the diffusion model. The dashed grey line denotes the actual motion direction, while the solid lines are laterally offset for clearer visualization.}
    \label{fig:diversity}
\end{figure}

Table~\ref{tab:seedsel} isolates the seed-selection module by fixing the controller and varying $\mathbf{q}_{0}^{*}$ across three sources: a manipulability seed, a joint-limit-centering seed, and the proposed DP seed. The two heuristics share one procedure, differing only in objective: null-space gradient ascent from $\mathbf{q}_{0}^{\mathrm{seed}}$ with periodic Newton re-projection onto the exact pose. As the ascent stays within the null space, both remain confined to the SMM branch of $\mathbf{q}_{0}^{\mathrm{seed}}$ and cannot cross to another, whereas DP samples a multi-modal distribution spanning multiple branches. Under every controller, DP leads both heuristics by $10$--$18$\,pp at comparable seed-generation cost ($\sim$45\,ms), confirming that the learned prior reaches a more favorable starting branch than branch-bound heuristic refinement. Fig.~\ref{fig:diversity} visualizes the multi-modal seeds the diffusion model samples on a representative task.

\begin{table*}[!htbp]
\centering
\setlength{\tabcolsep}{6pt}
\caption{Performance of the Proposed Initial Configuration Selection Method Compared with Heuristic Designs.}
\label{tab:seedsel}
\begin{threeparttable}
\begin{tabular}{llccc}
\toprule
$\pi$ & $\textbf{q}_{0}^*$ 
& $t$ (ms)
& \shortstack{Path Length (m)\\ {\scriptsize Mean / Std / Min / Max}} 
& \shortstack{Ratio to Reference ($\%$)\\ {\scriptsize Mean / Std / Min / Max}} \\
\midrule
\multirow{3}{*}{Classical}
& $\mathbf{q}_{\mu}$         & 47.40 & 0.36 / 0.16 / 0.01 / 1.14 & 61.3 / 28.8 / 0.8 / \phantom{0}555.4 \\
& $\mathbf{q}_{\mathrm{jl}}$  & 44.10 & \pinkbox{0.34} / 0.18 / 0.00 / 1.14 & \pinkbox{58.6} / 30.4 / 0.0 / \phantom{0}404.6 \\
& DP           & 45.20 & \limebox{0.45} / 0.18 / 0.01 / 1.25 & \limebox{76.1} / 26.9 / 0.8 / \phantom{0}601.3 \\
\midrule
\multirow{3}{*}{RL}
& $\mathbf{q}_{\mu}$ & 47.40 & \pinkbox{0.45} / 0.26 / 0.01 / 1.50 & \pinkbox{73.0} / 33.7 / 0.8 / 1058.3 \\
& $\mathbf{q}_{\mathrm{jl}}$              & 44.10 & 0.46 / 0.26 / 0.00 / 1.68 & 74.1 / 36.6 / 0.0 / \phantom{0}928.0 \\
& DP             & 45.20 & \limebox{0.53} / 0.26 / 0.01 / 1.45 & \limebox{85.0} / 28.8 / 0.8 / 1017.6 \\
\midrule
\multirow{3}{*}{Hybrid}
& $\mathbf{q}_{\mu}$         & 47.40 & 0.48 / 0.24 / 0.01 / 1.50 & 77.6 / 29.4 / 0.8 / \phantom{0}602.2 \\
& $\mathbf{q}_{\mathrm{jl}}$  & 44.10 & \pinkbox{0.48} / 0.26 / 0.00 / 1.68 & \pinkbox{77.1} / 35.0 / 0.0 / \phantom{0}928.0 \\
& DP           & 45.20 & \limebox{0.57} / 0.25 / 0.01 / 1.45 & \limebox{90.6} / 23.1 / 0.8 / 1000.8 \\
\bottomrule
\end{tabular}
\begin{tablenotes}
    \item[Note 1] $t$ is per-task seed-generation time.
    \item[Note 2] $\mathbf{q}_{\mu}$: manipulability-maximizing seed; $\mathbf{q}_{\mathrm{jl}}$: joint-limit-centering seed.
\end{tablenotes}
\end{threeparttable}
\end{table*}

\subsection{Ablation Study}

Unless otherwise specified, each ablation study uses the diffusion-based initial configuration under the hybrid controller with the default parameters introduced at the beginning of the experiment section.

\subsubsection{Classical Null-Space Controller Hyperparameters}

Table~\ref{tab:ablate_classical} isolates the three secondary objectives of Eq.~\eqref{eq:secondary} and then sweeps their weighted combination.  The full sweep covers a $4\times3\times3$ grid of $(k_{\mu},k_{\mathrm{jl}},k_{\theta})$ plus three single-objective settings ($39$ in total), of which the table lists a representative subset. We use the task's generating configuration $\mathbf{q}_{0}^{\mathrm{seed}}$ as the initialization for simplicity. The first three rows activate one objective at a time (gain $1.0$); the next rows increase $k_{\mu}$ at near-fixed $k_{\mathrm{jl}}$. Single objectives are uniformly weak ($58.7$--$62.3\%$), and any mixed combination surpasses them, indicating that the secondary objectives are complementary rather than redundant. Nonetheless, while tuned mixed-weight settings may help on individual tasks, no combination yields a meaningful improvement across the full evaluation set.

\begin{table*}[!htbp]
\centering
\setlength{\tabcolsep}{8pt}
\caption{Ablation of the Classical Secondary-Objective Weights.}
\label{tab:ablate_classical}
\begin{threeparttable}
\begin{tabular}{ccccc}
\toprule
$k_{\mu}$ & $k_{\mathrm{jl}}$ & $k_{\theta}$
& \shortstack{Path Length (m)\\ {\scriptsize Mean / Std / Min / Max}}
& \shortstack{Ratio to Reference ($\%$)\\ {\scriptsize Mean / Std / Min / Max}} \\
\midrule
1.0 & ---        & ---        & 0.36 / 0.15 / 0.01 / 1.15 & 62.3 / 29.5 / 0.8 / 1027.8 \\
---        & \pinkbox{1.0} & ---        & \pinkbox{0.34} / 0.17 / 0.02 / 1.28 & \pinkbox{58.7} / 28.5 / 3.5 / \phantom{0}521.9 \\
---        & ---        & 1.0 & 0.35 / 0.13 / 0.01 / 1.14 & 60.6 / 28.0 / 0.9 / \phantom{0}945.8 \\
\midrule
0.2 & 0.1 & 0.4 & 0.36 / 0.13 / 0.01 / 1.12 & 62.6 / 26.6 / 1.9 / \phantom{0}675.8 \\
0.4 & 0.2 & 0.4 & 0.37 / 0.14 / 0.02 / 1.16 & 63.7 / 26.7 / 2.5 / \phantom{0}675.9 \\
0.6 & 0.2 & 0.4 & 0.37 / 0.14 / 0.03 / 1.17 & 64.2 / 28.4 / 3.7 / \phantom{0}985.9 \\
0.8 & 0.2 & 0.4 & 0.37 / 0.15 / 0.03 / 1.17 & \limebox{64.5} / 28.8 / 3.2 / \phantom{0}986.3 \\
0.8 & 0.4 & 0.2 & 0.37 / 0.15 / 0.04 / 1.14 & 64.2 / 27.1 / 7.0 / \phantom{0}651.0 \\
\bottomrule
\end{tabular}
\begin{tablenotes}
    \item[Note 1] The cone gate $g(\theta)$ is rarely activated due to the conservativeness of the classical controller, so varying $k_{\theta}$ has little effect on performance.
\end{tablenotes}
\end{threeparttable}
\end{table*}

\subsubsection{Classifier-Free Guidance Scale}

Table~\ref{tab:ablate_cfg} varies the guidance scale $w$ of Eq.~\eqref{eq:cfg}, where $w=0$ ignores the task condition and larger $w$ amplifies it. The results show that performance peaks at $w=1.5$ ($90.6\%$), degrading under both unconditional sampling ($w=0$, $80.8\%$) and over-guidance ($w=3.0$, $84.2\%$). The seed-generation time $t$ is likewise lowest near $w=1.5$ and rises sharply at both ends ($133$\,ms at $w=0$, $94$\,ms at $w=3.0$, versus $\sim$46\,ms), as poorly conditioned samples need more sampling-and-projection retries to reach an IK-valid pose.

\begin{table*}[!htbp]
\centering
\setlength{\tabcolsep}{8pt}
\caption{Effect of the Guidance Scale $w$ on Path Length and Seed-Generation Cost.}
\label{tab:ablate_cfg}
\begin{threeparttable}
\begin{tabular}{lccc}
\toprule
$w$
& $t$ (ms)
& \shortstack{Path Length (m)\\ {\scriptsize Mean / Std / Min / Max}}
& \shortstack{Ratio to Reference ($\%$)\\ {\scriptsize Mean / Std / Min / Max}} \\
\midrule
0.0 & \pinkbox{133.50} & \pinkbox{0.50} / 0.25 / 0.01 / 1.49 & \pinkbox{80.8} / 29.3 / 0.8 / \phantom{0}566.5 \\
1.0 & \phantom{0}46.90 & 0.56 / 0.25 / 0.01 / 1.48 & 89.4 / 22.9 / 0.7 / \phantom{0}562.7 \\
1.5 & \phantom{0}\limebox{46.40} & \limebox{0.57} / 0.25 / 0.01 / 1.45 & \limebox{90.6} / 23.1 / 0.8 / 1000.8 \\
2.0 & \phantom{0}50.20 & 0.56 / 0.25 / 0.01 / 1.45 & 89.5 / 23.3 / 0.8 / \phantom{0}666.4 \\
3.0 & \phantom{0}94.00 & 0.52 / 0.24 / 0.01 / 1.40 & 84.2 / 26.2 / 0.8 / \phantom{0}506.0 \\
\bottomrule
\end{tabular}
\end{threeparttable}
\end{table*}

\subsubsection{Number of Diffusion Sampling Steps}

Table~\ref{tab:ablate_steps} examines the influence of the DDIM sampling steps on both time cost and seed quality. The results indicate that both metrics remain flat across the step range $10$--$100$. We therefore conclude that the result is essentially insensitive to the number of sampling steps, and we adopt $50$ as the default. The underlying reason may be that the seed quality is retained within a few denoising steps, as it largely depends on the SMM branch. Moreover, regarding time cost, although more sampling steps increase the sampling time, the dominant bottleneck of seed generation is the number of re-projection retries, which also depends on the seed quality.

\begin{table*}[!htbp]
\centering
\setlength{\tabcolsep}{6pt}
\caption{Effect of the DDIM Sampling Steps on Seed Quality and Time Cost.}
\label{tab:ablate_steps}
\begin{threeparttable}
\begin{tabular}{lccc}
\toprule
DDIM Steps
& $t$ (ms)
& \shortstack{Path Length (m)\\ {\scriptsize Mean / Std / Min / Max}}
& \shortstack{Ratio to Reference ($\%$)\\ {\scriptsize Mean / Std / Min / Max}} \\
\midrule
\phantom{0}10  & \pinkbox{48.80} & \limebox{0.572} / 0.245 / 0.010 / 1.448 & \limebox{90.6} / 20.9 / 0.9 / \phantom{0}659.0 \\
\phantom{0}20  & \limebox{46.40} & 0.570 / 0.244 / 0.010 / 1.487 & 90.5 / 22.1 / 0.8 / \phantom{0}683.3 \\
\phantom{0}50  & 46.50 & 0.571 / 0.246 / 0.010 / 1.448 & 90.6 / 23.1 / 0.8 / 1000.8 \\
100 & 47.30 & \pinkbox{0.569} / 0.247 / 0.010 / 1.590 & \pinkbox{90.2} / 21.9 / 0.8 / \phantom{0}567.4 \\
\bottomrule
\end{tabular}
\end{threeparttable}
\end{table*}

\subsubsection{Hybrid Switching Hyperparameters}

Table~\ref{tab:ablate_switch} varies the switching thresholds of Eq.~\eqref{eq:rho}. The top rows sweep a single threshold $\tau_{\mathrm{enter}}=\tau_{\mathrm{exit}}\in\{0.85,0.90,0.95,\\0.97,0.98,0.99\}$, tracing the performance comparison as the boundary region widens or shrinks. The bottom rows, in contrast, introduce a hysteresis gap with $\tau_{\mathrm{enter}}>\tau_{\mathrm{exit}}$, and the \emph{switches} column reports the mean number of transitions per episode. Results show that introducing the hysteresis gap largely reduces the number of switches compared with the single-threshold setting. We therefore adopt $(0.98,0.94)$ as the default, as it achieves a balanced performance between path-following length and switching stability.

\begin{table*}[!htbp]
\centering
\setlength{\tabcolsep}{4.8pt}
\caption{Ablation of the Switching Thresholds $(\tau_{\mathrm{enter}},\tau_{\mathrm{exit}})$.}
\label{tab:ablate_switch}
\begin{threeparttable}
\begin{tabular}{lccc}
\toprule
$(\tau_{\mathrm{enter}},\tau_{\mathrm{exit}})$
& \shortstack{Switches\\ {\scriptsize Mean / Episode}}
& \shortstack{Path Length (m)\\ {\scriptsize Mean / Std / Min / Max}}
& \shortstack{Ratio to Reference ($\%$)\\ {\scriptsize Mean / Std / Min / Max}} \\
\midrule
$(0.85,\,0.85)$ & \pinkbox{2.50} & \pinkbox{0.534} / 0.231 / 0.010 / 1.427 & \pinkbox{86.1} / 24.3 / 0.8 / \phantom{0}601.3 \\
$(0.90,\,0.90)$ & 2.49 & 0.556 / 0.239 / 0.010 / 1.437 & 88.8 / 22.9 / 0.8 / \phantom{0}601.3 \\
$(0.95,\,0.95)$ & 1.97 & 0.571 / 0.244 / 0.010 / 1.437 & 90.7 / 23.7 / 0.8 / 1001.2 \\
$(0.97,\,0.97)$ & 1.65 & \limebox{0.573} / 0.246 / 0.010 / 1.448 & \limebox{90.9} / 23.1 / 0.8 / 1001.9 \\
$(0.98,\,0.98)$ & 1.43 & 0.572 / 0.247 / 0.010 / 1.448 & 90.7 / 22.7 / 0.8 / 1002.2 \\
$(0.99,\,0.99)$ & 0.99 & 0.563 / 0.253 / 0.010 / 1.448 & 89.3 / 24.8 / 0.8 / 1017.6 \\
\midrule
$(0.98,\,0.94)$ & 0.44 & 0.571 / 0.246 / 0.010 / 1.448 & 90.6 / 23.1 / 0.8 / 1000.8 \\
$(0.99,\,0.93)$ & \limebox{0.32} & 0.567 / 0.247 / 0.010 / 1.448 & 90.0 / 23.9 / 0.8 / 1017.6 \\
\bottomrule
\end{tabular}
\begin{tablenotes}
    \item[Note 1] Only representative swept hysteresis combinations are listed.
\end{tablenotes}
\end{threeparttable}
\end{table*}

\section{Conclusions and Future Work}
\label{sec:conclusion}

In this paper, we developed a framework for extreme motion generation that advances a fixed-base manipulator along a prescribed straight line as far as possible. The framework adopts a step-level hybrid null-space controller, which delegates the well-sampled interior to an RL policy for length maximization and the near-boundary region to a robust classical controller through a hysteresis switching rule. Prior to execution, a conditional diffusion model initializes the manipulator on a favorable branch of the SMM. Experimental results demonstrated a longer path-following length than both the classical and learning-based controllers and showed that the hysteresis switching rule achieves a favorable balance between length and switching stability. Since our controller works entirely in the null space, it can naturally incorporate additional customized task constraints into the extreme motion process, similar to conventional null-space methods. By deferring contact with the kinematic boundaries that would otherwise terminate the motion prematurely, the framework enlarges the effective working range without scaling up the hardware or relocating the base. We are interested in exploring this direction further in future work.


%
%
%




\bibliographystyle{unsrt}
\bibliography{references}

\end{document}